# Scaling up Dynamic Topic Models

Arnab Bhadury[†][*], Jianfei Chen[†], Jun Zhu[†], Shixia Liu[‡]
[†]Department of Computer Science & Technology; State Key Lab of Intelligent Technology & Systems
[†]Tsinghua National Lab of Information Science & Technology; Center for Bio-Inspired Computing Research
[‡]School of Software, Tsinghua University, Beijing, 100084 China
abhadury@flipboard.com; chenjian14@mails.tsinghua.edu.cn; {dcszj, shixia}@tsinghua.edu.cn

## ABSTRACT

Dynamic topic models (DTMs) are very effective in discovering topics and capturing their evolution trends in time series data. To do posterior inference of DTMs, existing methods are all batch algorithms that scan the full dataset before each update of the model and make inexact variational approximations with mean-field assumptions. Due to a lack of a more scalable inference algorithm, despite the usefulness, DTMs have not captured large topic dynamics.

This paper fills this research void, and presents a fast and parallelizable inference algorithm using Gibbs Sampling with Stochastic Gradient Langevin Dynamics that does not make any unwarranted assumptions. We also present a Metropolis-Hastings based $O(1)$ sampler for topic assignments for each word token. In a distributed environment, our algorithm requires very little communication between workers during sampling (almost embarrassingly parallel) and scales up to large-scale applications. We are able to learn the largest Dynamic Topic Model to our knowledge, and learned the dynamics of 1,000 topics from 2.6 million documents in less than half an hour, and our empirical results show that our algorithm is not only orders of magnitude faster than the baselines but also achieves lower perplexity.

## 1. INTRODUCTION

Surrounded by data, statistical topic models have become some of the most useful machine learning tools to automatically analyze large sets of categorical data, including both text documents and images under some bag-of-words representations. Topic models can capture thematic structure that exists within a data corpus and finds a low dimensional representation of the documents. Such topical representations can be used for subsequent analysis tasks, such as clustering (29), classification (33; 34), and data visualization (16). One of the most popular topic models, Latent Dirichlet Allocation (LDA) (5), has seen large amounts of application in both industry (25) and academia. Since exact posterior inference of LDA is intractable, recent research has focused on speeding up the approximate inference methods for LDA from various directions, including stochastic/online inference (19), fast sampling algorithms (31; 7), and scalable systems (1).

While LDA is extremely useful, it has many simplistic assumptions that fail to capture some complicated structures underlying a large data corpus, such as the correlationship between multiple topics and the temporal evolution of topics in data streams. Correlated Topic Model (CTM) (3) is one such extension to LDA that introduces non-conjugate Logistic-Normal parameters to capture the correlation among topics. Though flexible in model capacity, the non-conjugacy makes approximating the posterior and scaling up a lot more difficult. Variational approximation was often adopted (3) under some unwarranted meanfield assumptions. The standard variational methods cannot deal with large datasets either. Recently, Chen et al. (8) scaled up the CTM using a novel (distributed) Gibbs Sampler with Data Augmentation, which does not make unnecessary mean-field assumptions; thereby leading to better performance in terms of both time efficiency and testing likelihood/perplexity.

Dynamic Topic Model (DTM) (4) is another extension to LDA that discovers topics and their evolution trends in time series data by chaining the time-specific topic-term distributions via a Markov process under a Logistic-Normal parameterization as in CTM. The non-conjugacy in the DTM model makes its large-scale posterior inference even more difficult and it remains a challenge in machine learning research. Existing inference algorithms of DTM have been focused on mean-field variational approximations, Laplace approximations or delta methods (23), which potentially lead to inaccurate results due to improper assumptions, require model specific derivations, and can only deal with small data corpora and learn a small number of topics. There has been a lot of recent research done to scale up variational inference (14; 6) to large data corpora but using variational methods, inferring the variational distribution over topics for a word in topic modeling is typically of $O(K)$ complexity, where $K$ is the number of topics, while recent sampling algorithms of LDA utilize the sparsity in the model (30) or use Metropolis-Hastings sampling with dedicated data structures (e.g., Alias tables (31; 15)) to incrementally bring the sampling complexity down to an amortized $O(1)$ per token. Our recent work (7) presents an even more efficient $O(1)$ sampler by optimizing the CPU cache access.

---

[*]Arnab Bhadury is now with Flipboard Inc. 210-128 W Hastings St. Vancouver, BC, V6B 1G8, Canada.

As shown by Wang et. al. (26) capturing a large number of topic trends in topic modeling is extremely important as it improves tasks such as advertisement and recommendations. Using variational approaches, it is difficult to support a large number of topics, while there also exist many machine learning applications where algorithms need to be fast while processing small scale data (17). To address these two different limitations, we present a novel scalable solution to do posterior inference using Gibbs Sampling which avoids any restricting assumptions and can be easily scaled up to capture thousands of topics from millions of documents in a parallel environment, while also being faster on single machines. To deal with the non-conjugacy in the model, within the Gibbs Sampling framework, we use recent developments in Stochastic MCMC and use Stochastic Gradient Langevin Dynamics (SGLD) (28) to sample the logistic normal parameters by observing only a mini-batch of the document set. Stochastic methods have shown to converge faster than their batch counterparts (19), and are particularly applicable to large-scale problems.

We use ideas from these state-of-the-art methods to derive a fast algorithm to capture topic trends of a large number of topics. We first derive the update equations for all the parameters, then show the algorithm's efficiency on a single multithreaded machine and further present a parallel algorithm to learn large Dynamic Topic Models on multiple machines. Our algorithm is very close to being "embarrassingly parallel"[1] and scales up extremely well with the number of time slices. We learn a large dynamic topic model from a 9 GB dataset consisting of 2.6 million documents in less than an hour. This is not only the biggest DTM but also the fastest inference algorithm to our knowledge.

In the rest of the paper, we first introduce some related work that inspires our research including LDA and its fast sampling algorithms, and the DTM model in Section 2. In Section 3, we introduce our proposed algorithm, and provide implementation details for multithreaded and distributed machines in Section 4. Experiments for both single machines and distributed machines are discussed in Section 5 and we conclude our paper in Section 6.

## 2. RELATED WORK

In this section, we briefly review some related work on the vanilla LDA and dynamic topic models.

### 2.1 Latent Dirichlet Allocation

Latent Dirichlet Allocation (LDA) is a probabilistic generative model of documents. It represents documents as an admixture of a set of $K$ topics to be learned from data. LDA uses conjugate Dirichlet-Multinomial parameters for both document-topic and topic-term distributions. Let $\mathcal{D} =$

---

[1]At the start of each iteration, there is a small synchronization step, but for the intensive computation part, there is no communication needed.

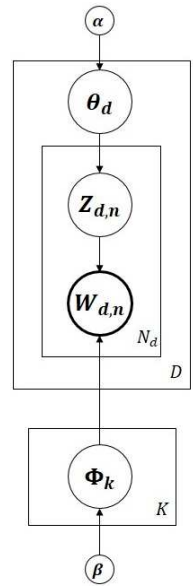

**Figure 1: Generative Process of Latent Dirichlet Allocation**

$\{W_d\}_{d=1}^D$ be a set of $D$ documents, where each document $W_d = \{W_{d,n}\}_{n=1}^{N_d}$ is a set of $N_d$ words $W_{d,n}$. The generating process, as shown in Figure 1, of LDA is

1. For each topic $k \in [K]$, draw $\boldsymbol{\Phi}_k \sim \text{Dir}(\boldsymbol{\Phi}_k|\boldsymbol{\beta})$;

2. For each document $d \in [D]$, draw the topic mixing proportion: $\boldsymbol{\theta}_d \sim \text{Dir}(\boldsymbol{\theta}_d|\boldsymbol{\alpha})$;

    (a) For each word $n \in [N_d]$, draw the topic assignment $Z_{d,n}$ and the word itself:

$$Z_{d,n} \sim \text{Mult}(Z_{d,n}|\boldsymbol{\theta}_d)$$

$$W_{d,n} \sim \text{Mult}(W_{d,n}|\boldsymbol{\Phi}_{Z_{d,n}}),$$

where $\boldsymbol{\alpha}$ and $\boldsymbol{\beta}$ are the Dirichlet parameters, $\boldsymbol{\Phi}_k$ is the term distribution of the topic $k$ of size $V$, $\boldsymbol{\theta}_d$ is the $K$-dimensional topic mixing distribution of document $d$, $\text{Dir}(\cdot)$ is the Dirichlet distribution, and $\text{Mult}(\cdot)$ is the Multinomial distribution.

Having conjugacy in the model allows the document-topic proportion $\boldsymbol{\Theta} := \{\boldsymbol{\theta}_d\}$ and the topic-term proportion $\boldsymbol{\Phi} := \{\boldsymbol{\Phi}_k\}$ to be integrated out analytically, yielding with a collapsed posterior distribution $p(\mathbf{Z}|\mathcal{D},\boldsymbol{\alpha},\boldsymbol{\beta})$, where $\mathbf{Z} := \{\mathbf{z}_d\}$ denotes the set of all topic assignments. Extensive research has been done to optimize the posterior inference of the collapsed model using both variational approximation (20) and sampling.

Recently, sampling based posterior inference algorithms have been paid close attention to because of their simplicity and the fact that they have shown to provide sparser results that can be further optimized. Once the conjugate parameters are integrated out from the joint distribution, a simple Gibbs Sampler can draw topic indices $Z_{d,n}$ by the following conditional probability:

$$p(Z_{d,n}|rest) \propto \frac{(C_k^{d,\neg(d,n)} + \alpha_k)(C_k^{w,\neg(d,n)} + \beta_w)}{C_k^{\neg(d,n)} + \bar{\beta}}. \quad (1)$$

Here, $C_k^d$ stands for the number of times that topic $k$ has been observed in document $d$, $C_k^w$ denotes the number of times a word $w$ has been observed as a topic $k$ in the corpus, $C_k$ stands for the number of times topic $k$ has been assigned to a word in the corpus, and $rest$ denotes all other variables except $Z_{d,n}$. The superscript $\neg(d,n)$ specifies the count ignoring the topic index for the word in $(d,n)$ position; and $\bar{\beta}$ is the sum of all the $\beta_w$s.

Eq. (1) is the original Collapsed Gibbs Sampler conditional proposed by Griffiths et al. (12) that requires an $O(K)$ time complexity to sample each token in the data. Yao et al. (30) rewrite Eq. (1) into a different form to take advantage of sparsity that exists within the model[2] to lower the time complexity of sampling each token down to $O(K_d + K_w)$, where $K_d$ denotes the number of topics that exist in a document $d$ and $K_w$ stands for the number of different topics a word $w$ has been observed by. For models with a large $K$, we often have $K_d \ll K$ and $K_w \ll K$.

AliasLDA (15) factorizes the same conditional in Eq. (1) into two components, a sparse component and a dense component. The sparse component (document-component) is sampled in a method analogous to SparseLDA (30) in $O(K_d)$, and for sampling the dense component, an Alias table (21) is created to generate $K$ stale samples and Metropolis-Hastings tests (18; 13) are used to accept parameter updates in amortized $O(1)$ time complexity.

LightLDA(31) builds on that by removing the sparse component altogether, and uses an Alias table to generate two proposals that are factors of the true conditional. LightLDA alternatively samples from the two high probable proposal distributions, and brings the amortized sampling complexity down by another magnitude to $O(1)$. WarpLDA (7) further improves the efficiency of LightLDA by optimizing the access of CPU cache, making the $O(1)$ algorithm more efficient in practice.

This research builds on their insights to create a fast sampler of topic indices for DTM.

## 2.2 Dynamic Topic Models

Dynamic Topic Model (DTM) (4) is a topic model that is used to model time series data. Since a Dirichlet distribution is not suitable to model state changes, DTM adopts a logistic-normal parameterization as in CTM (3), chains the parameters together in a Markovian structure, and allows evolution of parameters with a Gaussian noise. The Gaussian variables are mapped to the simplex from where the Multinomial variables are drawn.

Given a set of data $\mathcal{D} = \{\mathcal{D}_t\}_{t=1}^T$, where $\mathcal{D}_t$ denotes the dataset at time slice $t$ and $T$ is the total number of time slices. For each time slice $t$, DTM first samples the evolved parameters from linear dynamic systems, namely,

$$\boldsymbol{\alpha}_t \sim \mathcal{N}(\boldsymbol{\alpha}_t|\boldsymbol{\alpha}_{t-1}, \sigma^2 I)$$

$$\forall k \in [K]: \quad \boldsymbol{\Phi}_{k,t} \sim \mathcal{N}(\boldsymbol{\Phi}_{k,t}|\boldsymbol{\Phi}_{k,t-1}, \beta^2 I),$$

where $\sigma$ and $\beta$ are variance parameters. Then, for a document $d$ at time $t$, the words are generated as follows:

$$\boldsymbol{\eta}_{d,t} \sim \mathcal{N}(\boldsymbol{\eta}_{d,t}|\boldsymbol{\alpha}_t, \psi^2 I),$$

---

[2]Assuming that a document $d$ only contains a subset of topics, and a word $w$ is only assigned to a subset of topics while sampling. Following this assumption, (1) could be multiplied for a sparser expression. The second part of this assumption does not necessarily hold in large data corpora.

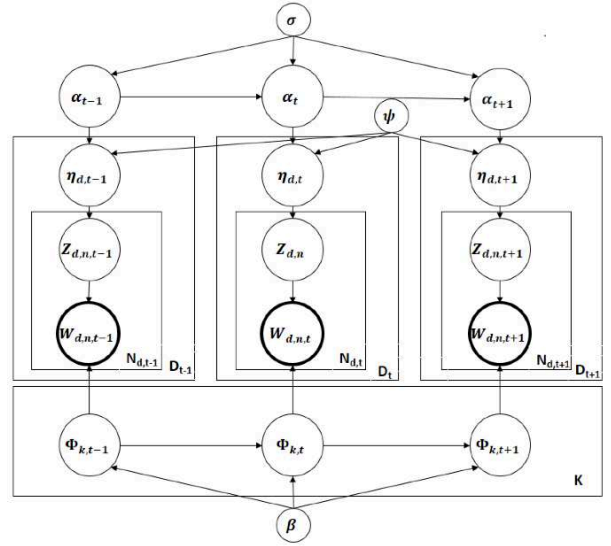

Figure 2: Generative Process of DTM

$$\forall \text{n} \in [N_{d,t}]: \quad Z_{d,n,t} \sim \text{Mult}(Z_{d,n,t}|\pi(\boldsymbol{\eta}_{d,t})),$$
$$\forall \text{n} \in [N_{d,t}]: \quad W_{d,n,t} \sim \text{Mult}(W_{d,n,t}|\pi(\boldsymbol{\Phi}_{Z_{d,n,t},t})),$$

where $\pi(\boldsymbol{\alpha}) = \boldsymbol{\gamma}$ with each element being $\gamma_k = \frac{\exp(\alpha_k)}{\sum_j \exp(\alpha_j)}$ under a soft-max transformation, and $\psi$ is another variance parameter.

After observing the data $\mathcal{D}$, the posterior distribution conditioned on the hyper-parameters is given by:

$$p(\boldsymbol{\alpha}, \boldsymbol{\eta}, \boldsymbol{\Phi}, \mathbf{Z}|\mathcal{D}, \sigma, \beta, \psi) \propto \prod_{t=1}^T \mathcal{N}(\boldsymbol{\alpha}_t|\boldsymbol{\alpha}_{t-1}, \sigma^2 I) \times$$
$$\prod_{k=1}^K \mathcal{N}(\boldsymbol{\Phi}_{k,t}|\boldsymbol{\Phi}_{k,t-1}, \beta^2 I) \prod_{d=1}^{D_t} \mathcal{N}(\boldsymbol{\eta}_{d,t}|\boldsymbol{\alpha}_t, \psi^2 I) \times$$
$$\prod_{n=1}^{N_{d,t}} \text{Mult}(Z_{d,n,t}|\pi(\boldsymbol{\eta}_{d,t})) \text{Mult}(W_{d,n,t}|\pi(\boldsymbol{\Phi}_{Z_{d,n,t},t})).$$

The non-conjugacy of Gaussian and multinomial variables makes exact inference of the model intractable. Unlike LDA, the topic-term proportion parameters $\boldsymbol{\Phi}_t$ and the document-topic proportion parameters $\boldsymbol{\eta}_t$ cannot be analytically integrated out, which makes the task of approximating the posterior more complicated.

Blei & Lafferty (4) propose a variational Kalman filtering approach to infer the posterior distribution by making an unwarranted mean-field approximation, which is not scalable to large datasets. Other fast variational methods (14; 6) could be generalized to do posterior inference but as discussed earlier, variational methods are not amenable to support a large number of topics due to the inability to explore model sparsity. In the following section, we present an even faster Gibbs Sampler that is efficient in capturing large dynamic topic models with many topics.

## 3. GIBBS SAMPLER FOR DTM

We now present our blockwise Gibbs Sampler for the posterior inference of DTM. Unlike LDA or CTM, we cannot integrate any of the parameters out because of the non-conjugacy from both directions, therefore we sample each parameter separately. The parameters that we need to infer are $\boldsymbol{\alpha}_t$, $\boldsymbol{\eta}_{d,t}$, $\boldsymbol{\Phi}_{k,t}$ and $\mathbf{Z}_{d,n,t}$. The following subsections show the derivation and the sampler update equations of each parameter conditioned on all the other parameters.

### 3.1 Sampling $\alpha_t$

It is easy to see from Figure 2 that $\boldsymbol{\alpha}_t$ has a Markovian structure and conditioned on its Markov Blanket (MB), we get:

$$p(\boldsymbol{\alpha}_t|\text{MB}(\boldsymbol{\alpha}_t)) \propto \mathcal{N}(\boldsymbol{\alpha}_t|\boldsymbol{\alpha}_{t-1}, \sigma^2 I)\mathcal{N}\left(\boldsymbol{\alpha}_{t+1}|\boldsymbol{\alpha}_t, \sigma^2 I\right) \\ \prod_{d=1}^{D_t} \mathcal{N}\left(\boldsymbol{\eta}_{d,t}|\boldsymbol{\alpha}_t, \psi^2 I\right). \quad (2)$$

With three Gaussian terms, we simply use the "completing the square" trick and get:

$$\boldsymbol{\alpha}_t \sim \mathcal{N}\left(\boldsymbol{\alpha}_t|\hat{\boldsymbol{\mu}}, \hat{\boldsymbol{\Lambda}}^{-1}\right), \quad (3)$$

where the mean and covaraiance matrix are evaluated as:

$$\hat{\boldsymbol{\mu}} = \overline{\boldsymbol{\alpha}}_t + \overline{\boldsymbol{\eta}}_{d,t} - \boldsymbol{\Lambda}^{-1}\left(\frac{2}{\sigma^2}\overline{\boldsymbol{\eta}}_{d,t} + \frac{D_t}{\psi^2}\overline{\boldsymbol{\alpha}}_t\right) \quad (4)$$

$$\overline{\boldsymbol{\eta}}_{d,t} = \frac{1}{D_t}\sum_{d=1}^{D_t}\boldsymbol{\eta}_{d,t}, \quad \overline{\boldsymbol{\alpha}}_t = \frac{\boldsymbol{\alpha}_{t+1} + \boldsymbol{\alpha}_{t-1}}{2}, \\ \hat{\boldsymbol{\Lambda}} = \left(\frac{2}{\sigma^2} + \frac{D_t}{\psi^2}\right)I. \quad (5)$$

The key thing to note here is that evaluating the mean $\hat{\boldsymbol{\mu}}$ is an $O(K)$ operation as long as we keep record of $\overline{\boldsymbol{\eta}}_{d,t}$ because $\hat{\boldsymbol{\Lambda}}$ is a diagonal matrix. Also note that $\hat{\boldsymbol{\Lambda}}$ is constant for all documents in the time slice $t$.

### 3.2 Sampling $\eta_{d,t}$

Conditioned on $\boldsymbol{\alpha}_t$ and $\mathbf{Z}_{d,t}$, the posterior conditional for $\boldsymbol{\eta}_{d,t}$ is:

$$p(\boldsymbol{\eta}_{d,t}|\boldsymbol{\alpha}_t, \mathbf{Z}_{d,t}) \propto \mathcal{N}(\boldsymbol{\eta}_{d,t}|\boldsymbol{\alpha}_t, \psi^2 I) \\ \times \prod_{n=1}^{N_{d,t}} \text{Mult}(Z_{d,n,t}|\pi(\boldsymbol{\eta}_{d,t})) \quad (6)$$

This is a multivariate Bayesian Logistic Regression model with a Gaussian prior and Multinomial observations $Z_{d,n,t}$. To infer the posterior, we use a novel method developed by Welling & Teh called Stochastic Gradient Langevin Dynamics (SGLD) (28).

SGLD is an iterative learning algorithm that uses mini-batches to perform updates, which is suitable for large datasets. It builds on the well-known optimization algorithm, Stochastic Gradient Descent, and adds some Gaussian noise to each update so that it can generate samples from the true posterior and not just collapse to the MAP/MLE solution. Let $p(\theta|\{x_n\}) \propto p(\theta)\prod_n p(x_n|\theta)$ be an arbitrary generative model with prior $p(\theta)$ and likelihood $p(x|\theta)$. The SGLD parameter update for $\theta$ at the $i^{th}$ iteration is:

$$\Delta\theta_i = \frac{\epsilon_i}{2}\left(\nabla \log p(\theta) + \frac{N}{M}\sum_{n=1}^{M}\nabla\log p(x_n^i|\theta)\right) + \xi_i, \quad (7)$$

$$\xi_i \sim \mathcal{N}(\xi_i|0, \epsilon_i), \quad (8)$$

where $N$ is the number of data points in the dataset, and $M$ is a mini-batch created from those $N$ data points. Welling & Teh show that a Metropolis-Hastings test is unnecessary if we update $\epsilon_i$ in a way such that as $i$ increases, $\epsilon_i \to 0$ and the discretization error of Langevin Dynamics becomes negligible bringing the MH rejection probability close to 0. In our research, we use $\epsilon_i = a \times (b+i)^{-c}$ as our heuristic, and it satisfies the aforementioned condition.

Taking the gradient of the natural logarithm of the posterior conditional of $\boldsymbol{\eta}_{d,t}$ in (6), we get:

$$\nabla_{\eta_{d,t}^k} \log p(\boldsymbol{\eta}_{d,t}|\boldsymbol{\alpha}_t, \mathbf{Z}_{d,t}) = -\frac{1}{\psi^2}(\eta_{d,t}^k - \alpha_t^k) \\ + \sum_{n=1}^{N_{d,t}}\left(\delta(Z_{d,n,t} = k) - \pi(\boldsymbol{\eta}_{d,t})_k\right), \quad (9)$$

where $\delta(Z_{d,n,t} = k)$ is the Kronecker delta function. On closer attention to the gradient, it is easy to see that the first term in the gradient of the likelihood term is simply the number of times the topic $k$ has been observed in document $d$ in time slice $t$, and hence, the summation in (9) can be replaced, and (9) can be simply rewritten as:

$$\nabla_{\eta_{d,t}^k} \log p(\boldsymbol{\eta}_{d,t}|\boldsymbol{\alpha}_t, \mathbf{Z}_{d,t}) = -\frac{1}{\psi^2}(\eta_{d,t}^k - \alpha_t^k) \\ + C_{d,t}^k - (N_{d,t} \times \pi(\boldsymbol{\eta}_{d,t})_k), \quad (10)$$

where $N_{d,t}$ is the number of words in document $d$ of time slice $t$ and $C_{d,t}^k$ stands for the number of times the topic $k$ has been observed in document $d$ in time slice $t$. This equation can be directly put into the SGLD update equation, and the count matrix can be updated while storing topic indices. Updating the topic proportion of each document takes $O(K)$ time, if we store the softmax normalization constant during the first evaluation.

### 3.3 Sampling $\Phi_{k,t}$

$\boldsymbol{\Phi}_k$s are linked together as a Markov chain similar to $\boldsymbol{\alpha}_t$s and the posterior conditioned on its Markov Blanket is:

$$p(\boldsymbol{\Phi}_{k,t}|\text{MB}(\boldsymbol{\Phi}_{k,t})) \propto \mathcal{N}(\boldsymbol{\Phi}_{k,t}|\boldsymbol{\Phi}_{k,t-1}, \beta^2 I) \\ \times \mathcal{N}(\boldsymbol{\Phi}_{k,t+1}|\boldsymbol{\Phi}_{k,t}, \beta^2 I) \\ \times \prod_{d=1}^{D_t}\prod_{n=1}^{N_{d,t}} \text{Mult}(W_{d,n,t}|\pi(\boldsymbol{\Phi}_{k,t})). \quad (11)$$

Since the first two terms are Gaussians with the same variance, they can be multiplied together to get a new Gaussian, and we can treat it as the prior. Using the "completing the square" trick again, Eq. (11) can be rewritten as:

$$p(\boldsymbol{\Phi}_{k,t}|\text{MB}(\boldsymbol{\Phi}_{k,t})) = \mathcal{N}(\boldsymbol{\Phi}_{k,t}|\overline{\boldsymbol{\Phi}}_{k,t}, \frac{\beta^2}{2}I) \times \\ \prod_{d=1}^{D_t}\prod_{n=1}^{N_{d,t}} \text{Mult}(W_{d,n,t}|\pi(\boldsymbol{\Phi}_{k,t})), \quad (12)$$

where

$$\overline{\boldsymbol{\Phi}}_{k,t} = \frac{\boldsymbol{\Phi}_{k,t+1} + \boldsymbol{\Phi}_{k,t-1}}{2} \quad (13)$$

Combining Eq. (11) with Eq. (12), we, again, have a Bayesian Logistic Regression model with a Gaussian prior and $W_{d,n,t}$ as multinomial observations. If we take the gradient of

the natural logarithm of this Bayesian Logistic Regression model, we get:

$$\nabla_{\Phi_{k,t}^w} \log p(\Phi_{k,t}| \text{ MB}(\Phi_{k,t})) = \frac{\Phi_{k,t+1}^w + \Phi_{k,t-1}^w - 2\Phi_{k,t}^w}{\beta^2} + C_{k,t}^w - (C_{k,t} \times \pi(\Phi_{k,t})_w), \quad (14)$$

where $C_{k,t}^w = \sum_{d=1}^{D_t} \sum_{n=1}^{N_{d,t}} (\delta(W_{d,n,t} = w, Z_{d,n,t} = k))$ and $C_{k,t} = \sum_{d=1}^{D_t} \sum_{n=1}^{N_{d,t}} (\delta(Z_{d,n,t} = k))$.

Using cached matrices, only the first and the third terms need to be evaluated, and $\Phi_t$ can be sampled in $O(VK)$ time using SGLD. Since we use a small batch of documents to sample $\Phi_t$, we clear $C_{k,t}^w$ and $C_{k,t}$ at the start of each iteration.

## 3.4 Sampling $Z_{d,n,t}$

From the generative process of DTM, as shown earlier in Figure 2, it can be found that given $\eta$ and $\Phi$, $Z$ is conditionally independent of the rest of the parameters in the model. Therefore, sampling the topic indices for DTM conditioned on the topic-term proportions and the topic-document proportions can be collapsed to:

$$p(Z_{d,n,t} = k|\eta_{d,t}^k, \Phi_{k,t}^w) \propto \underbrace{\exp(\eta_{d,t}^k)}_{doc} \underbrace{\exp(\Phi_{k,t}^w)}_{word} \quad (15)$$

However, since sampling a token requires evaluating the normalization constant, the naive method requires $O(K)$ time to sample each token, and for fast inference, it is absolutely vital to have a fast sampler for $Z$ since it is the most expensive parameter to sample in DTM. We take ideas from recent advances in the inference of Latent Dirichlet Allocation to optimize the sampling, where authors use Alias Tables (as mentioned in Section 2) to generate $K$ samples in $O(K)$ time that are accepted using a Metropolis-Hastings (MH) Test, thus bringing the amortized sampling complexity to $O(1)$(31; 15).

Alias sampling (21) takes advantage of the fact that while generating one sample from a $K$ dimensional vector is of $O(K)$ complexity, but subsequent samples only require an $O(1)$ time. It transforms the multinomial sampling problem into an easy uniform sampling one, which requires a simple lookup on the "Alias tables", as shown in Figure 3.

But since Eq. (15) is not a static distribution, a Metropolis-Hastings (MH) Test is done to check whether the new sample should be accepted or rejected. While in theory, an $O(1)$ complexity is great, but in practice, we desire high acceptance of the stale samples from the MH Test.

The key to high acceptance rates in MH tests is to generate good proposals, and similar to Yuan et al. (31), we break Eq. (15) into two products $\exp(\eta_{d,t}^k)$ and $\exp(\Phi_{k,t}^w)$, and use one of them alternatively to generate proposals. We call proposals generated from the topic-term parameter the *word-proposal* and proposals generated from the topic-document parameter the *doc-proposal*.

To generate both proposals, we generate Alias tables, and generate $K$ samples, and use them until we run out of generated samples. When we do not have any more samples, we build a new Alias table, and generate new samples again. The doc-proposal $p_d(k)$ and its acceptance probability $A_d$ for topic $k$ are given by:

$$p_d(k) \propto \exp(\eta_{d,t}^k) \quad (16)$$

| Parameter | Sampling Complexity |
|---|---|
| $\alpha$ | $O(K)$ |
| $\eta$ | $O(D_m K)$ |
| $\Phi$ | $O(VK)$ |
| $Z$ | $O(D_m N_d)$ |

Table 1: Sampling complexities of each parameter in DTM each iteration. $D_m$: Size of the mini-batch, $N_d$: Number of words in a sample document d.

$$\begin{aligned} A_d &= \min\left(1, \frac{\exp(\eta_{d,t}^k + \Phi_{k,t}^w + \eta_{d,t}^s)}{\exp(\eta_{d,t}^s + \Phi_{s,t}^w + \eta_{d,t}^k)}\right) \\ &= \min\left(1, \frac{\exp(\Phi_{k,t}^w)}{\exp(\Phi_{s,t}^w)}\right). \end{aligned} \quad (17)$$

Analogous to the word proposal, the word-proposal $p_w(k)$ and its acceptance probability $A_w$ for topic $k$ are given by:

$$p_w(k) \propto \exp(\Phi_{k,t}^w) \quad (18)$$

$$\begin{aligned} A_w &= \min\left(1, \frac{\exp(\eta_{d,t}^k + \Phi_{k,t}^w + \Phi_{s,t}^w)}{\exp(\eta_{s,t}^k + \Phi_{s,t}^w + \Phi_{k,t}^w)}\right) \\ &= \min\left(1, \frac{\exp(\eta_{d,t}^k)}{\exp(\eta_{d,t}^s)}\right). \end{aligned} \quad (19)$$

It is easy to see that $A_w$ is high when the proposed topic $k$ is commonly seen in document $d$ and $A_d$ is high when the word $w$ is morse often observed as a topic $k$.

The sampling complexities of each parameter have been summarized in Table 1, and in a multicore environment, all of the parameters could be sampled completely independently after updating the count matrices at the start of each iteration.

## 4. MULTITHREADED AND DISTRIBUTED IMPLEMENTATION

In this section we present a distributed implementation of our algorithm, which utilizes two levels of parallelism: multi-machines and multi-threads.

DTM, itself, has an "embarrassingly parallel" structure and can be exploited by assigning each time slice to a separate worker. The data of different time slices can be stored in different machines without ever needing to move them around. We implement DTM in parallel using Message-Passing Interface (MPI) and pass the required data to neighbouring workers at the start of each iteration. The Markovian structure in the model requires $\alpha_t$ and $\Phi_t$ to be passed to adjacent nodes, and after that no further communication is required among these nodes.

Within each worker process, there is an additional level of multi-thread parallelism, implemented using C++11 `std::thread` in our system. We create three threads to sample $\eta, Z, \Phi$ respectively for the time slice assigned to a worker. We relax the original blockwise Gibbs sampling by adopting a Jacobi style iteration. Instead of using the most recent value, we sample $\eta$ conditioned on $Z$ and $\Phi$ from the previous iteration and likewise for $Z$ and $\Phi$. We can therefore sample $\eta, Z$ and $\Phi$ independently. Better implementation with no Jacobi relaxations and better parallelism can be designed, which sample $\eta, Z$ and $\Phi$ sequentially and partition computational tasks by d for $Z, \eta$ and by k for $\Phi$.

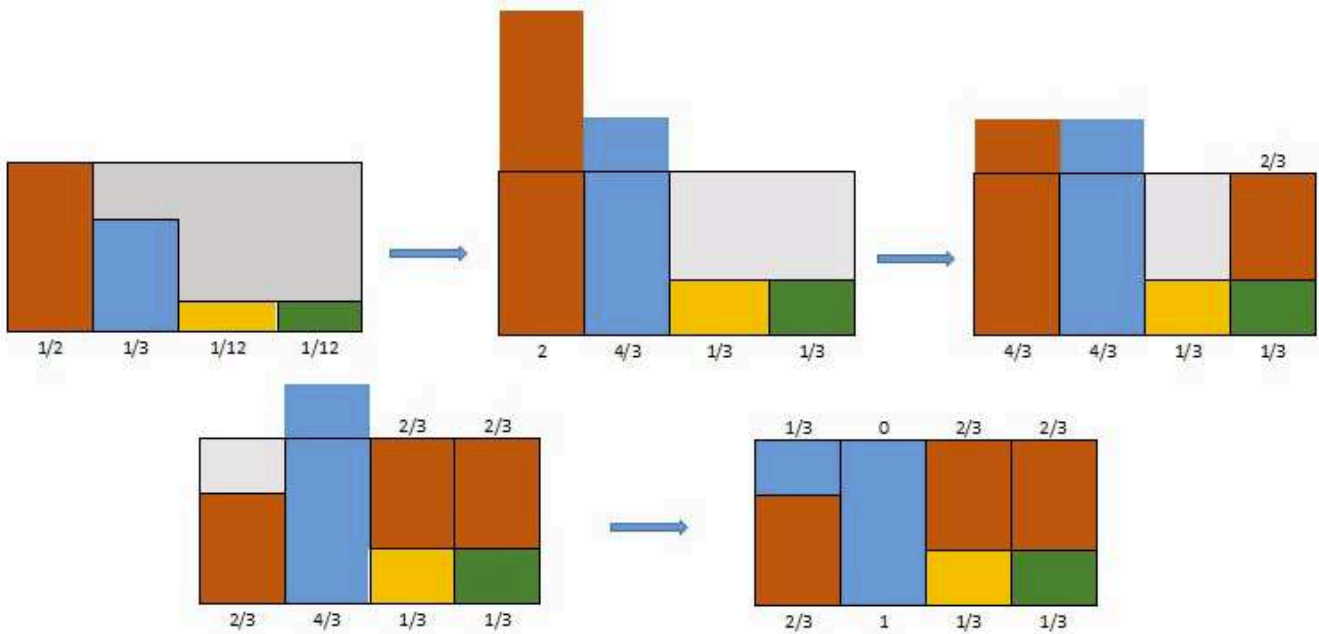

Figure 3: An example of how alias table (21) is constructed for a four dimensional non-uniform vector. The figure shows how a non-uniform vector of different probabilities can be compactly stored in a table with the constraint that each column indexes a maximum of two elements in the vector. Using an alias table, generating samples samples only requires two uniform samples; first to pick the column, and then doing a weighted coin-toss to sample an element. Constructing Alias Tables takes $O(K)$ time, where $K$ is the number of dimensions in the vector.

We adopt the aforementioned 3-thread approach purely for ease of implementation.

While using distributed computing, we often come across huge datasets (Bing News, in Section 5), where after trimming of vocabulary, the number of documents is much larger than the size of the vocabulary. In this case, one option is to use a mini-batch size ($D_m$) such that $D_m \times N_{avg} \approx V \times K$ for load balancing between threads, where $N_{avg}$ denotes the average length of documents. Another option is to divide the sampling of **Z** and $\boldsymbol{\eta}$ into more cores than for $\boldsymbol{\Phi}$.

## 5. EXPERIMENTS

We now present our algorithm in two experiment platforms, the first being a single machine in a multithreaded environment, and the second being a 6-node cluster consisting of 72 cores in total. Our results demonstrate that our algorithm (GS-SGLD) is substantially faster than the existing baseline (VKF) on both single machine and parallel environments. We also capture the largest Dynamic Topic Model to our knowledge.

### 5.1 Datasets and Setups

There are two datasets used for the experiments. The first consists of NIPS full papers from the year 1987 to 1999[3], where we pick the most frequent 8,000 words as vocabulary with stop words removed. We divide the dataset into thirteen time slices by year. We also use a large dataset that consists of all the news containing the word "Obama" from Bing News during the years from 2012 to 2015. This second dataset consists of more than 2.6 million documents, and are divided into 29 time slices, according to months. The vocabulary is trimmed down to 15,000 words after removing stop words, and looking at the most frequent words.

We partition each time slice of both datasets to contain a training set and a testing set with observed and held-out parts to evaluate perplexity. We use the partially observed document method (22).

For the NIPS dataset, we run our GS-SGLD sampler on a single machine, and Figure 4 shows the evolution of one of the 50 topics captured. The figure shows different methods and algorithms used in Computer Vision over the years, and the increase in the problem complexity as algorithms got better (Face Recognition, Object Recognition).

In the Bing News dataset, we capture many political trends, and Figure 6 shows one such evolution of the topic regarding the possibility of Syria possessing mass-destruction weapons. This trend shows an increasing involvement of Russia and the United States on the issue, and words such as "agreement" are captured in September 2014, when agreements were announced to eliminate chemical weapon stockpiles. The trend captured is in complete accordance with the Wikipedia article on the issue.[4]

### 5.2 Single Machine Experiments

In this section, we compare our inference algorithm to the existing variational Kalman filtering (VKF) method (4) for estimating DTM's model parameters.

---

[3]Available here: `http://www.cs.nyu.edu/~roweis/data.html`

[4]Wikipedia link on the issue: `http://en.wikipedia.org/wiki/Syria_and_weapons_of_mass_destruction`

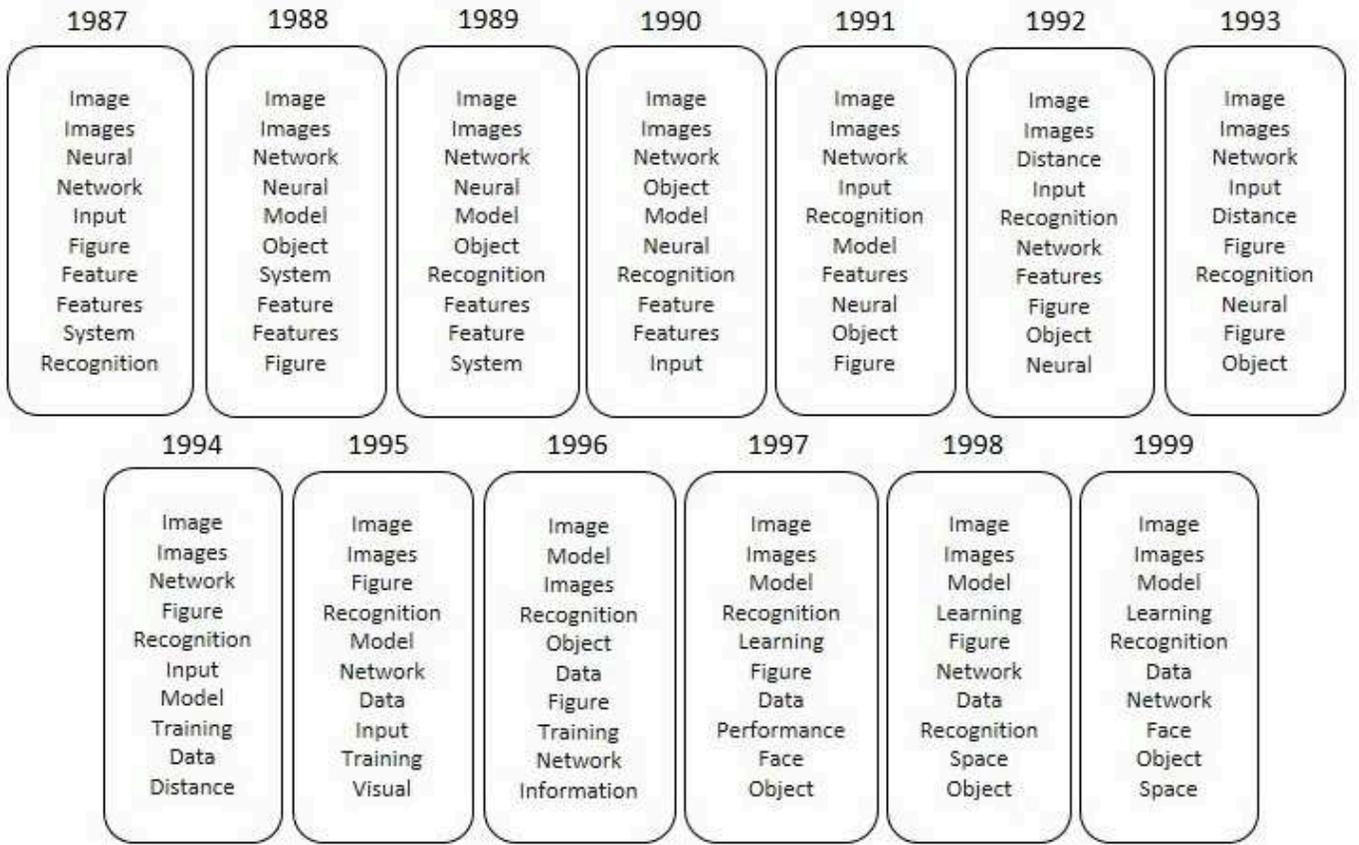

Figure 4: Evolution of Image Processing in NIPS from 1987-1999.

We run our GS-SGLD algorithm and the VKF algorithm on the NIPS dataset containing a total of 1,740 documents divided in 13 time slices. We set our SGLD learning rate to $\epsilon_i = 0.5 \times (100 + i)^{-0.8}$ and use a mini-batch size of 60 documents. Figure 5 shows our results compared to the baseline when we capture 50 topics in the dataset, and in Figure 7, we further show that the perplexity of each time slice decreases at approximately the same rate, which is also a desirable property for our algorithm. For clarity, we only plot the first five time slices and the average perplexity of 13 time slices.

We find that our algorithm (GS-SGLD) is significantly faster than the variational Kalman filtering (VKF) approach because of using a stochastic algorithm and the amortized $O(1)$ sampler for the topics for each token. In addition, GS-SGLD also achieves a slightly lower perplexity, compared to VKF, even after running the algorithm until both algorithms converge due to not making any unwarranted mean-field assumptions.

### 5.3 Parallel Experiments

In this section, we show the scalability of our algorithm by inferring $1,000$ topics from the Bing News dataset described earlier. The dataset consists of news of 29 time slices, and we use 58 cores for our experiment, to infer a large Dynamic Topic Model. We use a mini-batch size of 8,000 and set our SGLD learning rate $\epsilon_i$ to $0.5 \times (1000 + i)^{-0.75}$. We also run our code on the NIPS dataset with the same parameters as the ones described in the Single Machine Experiments section. The results are summarized in Table 2.

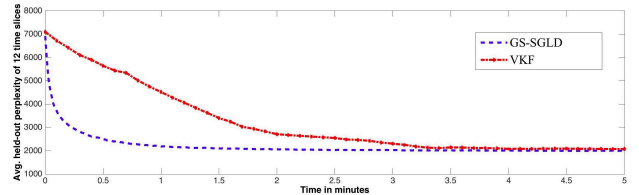

Figure 5: Perplexity comparison of GS-SGLD with VKF over time in seconds.

| Dataset | Docs | Time Slices | Topics | Run. Time |
|---------|------|-------------|--------|-----------|
| NIPS | 1740 | 13 | 50 | 118.72 |
| Bing news | 2.6M | 29 | 1000 | 1694.83 |

Table 2: Running time comparison (in seconds) of two datasets in a parallel environment.

We further show that our algorithm scales up well with an increasing amount of time slices due to its embarrassingly parallel nature. Assuming we have more cores to work with than time slices, there is a little communication overhead added to the running time, but the sampling complexity remains constant. The experiment in Table 3 uses the same dataset but we trim each time slice's documents to 50K so that we can efficiently run the algorithm on a single machine as well. During this experiment, we choose our mini-batch size to be 2000 and we iterate until convergence (around 60 iterations). The scalability is graphed in Figure 8.

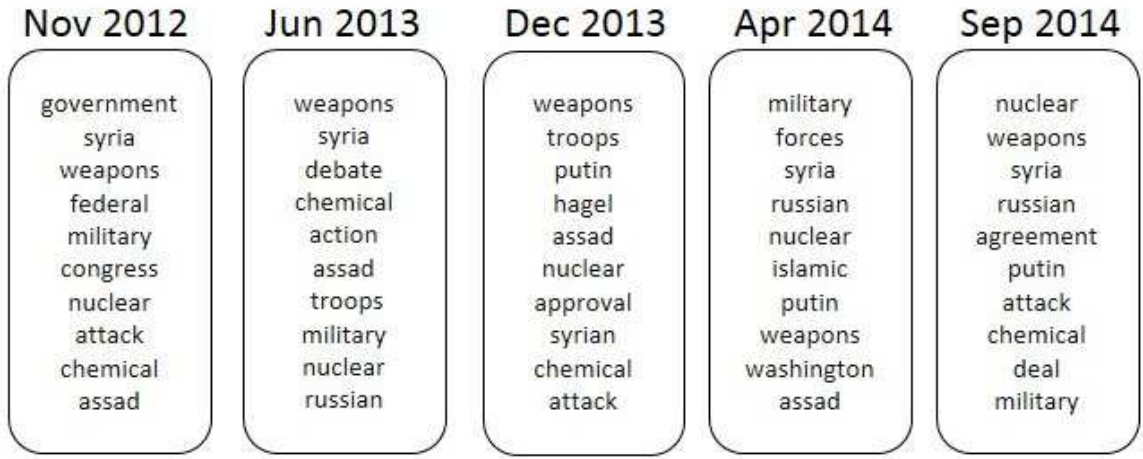

Figure 6: Evolution of Syrian possible threat of possessing mass-destruction weapons.

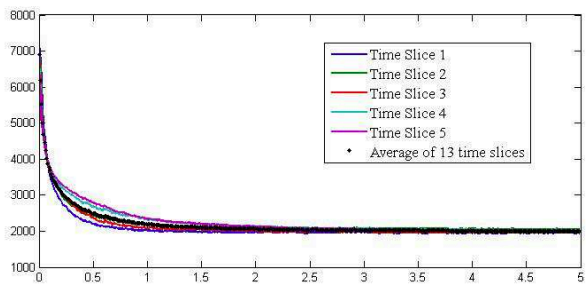

Figure 7: Perplexity comparison of different time slices.

| Num. of Time Slices | S-M GS-SGLD | P GS-SGLD |
|---|---|---|
| 5 | 712.81 (2) | 208.74 (10) |
| 8 | 1106.25 (2) | 227.38 (16) |
| 11 | 1577.83 (2) | 296.51 (22) |
| 15 | 2200.01 (2) | 302.97 (30) |
| 20 | N/A* (2) | 357.11 (40) |
| 29 | N/A* (2) | 398.32 (58) |

Table 3: Running time comparison of GS-SGLD inference in single machine and parallel settings in seconds. *not finished in one hour. (number) specifies the number of cores.

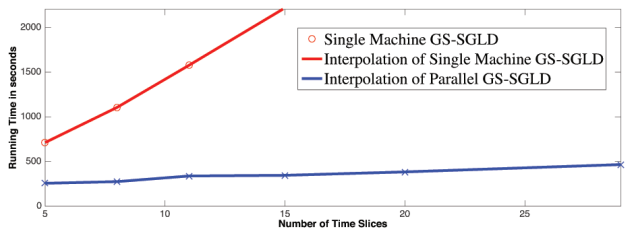

Figure 8: Running Time comparison of single machine and parallel GS-SGLD inference.

## 6. DISCUSSIONS AND FUTURE WORK

It is also worth mentioning that there are other dynamic topic models that can also capture topic trends over time such as Dynamic Mixture Model (DMM) (27) and Topics over Time (ToT) (24) but the Dynamic Topic Model proposed by Blei et. al has the distinct advantage of being able to naturally capture the correlation among topics over time by allowing the covariance of $\eta$ to be non-diagonal matrices. Relaxing this condition only requires a small update in the $\eta$ and $\alpha$ samplers, and hence DTM is especially appealing to us. Furthermore, DMM and ToT can be parallelized using the existing parallel frameworks of LDA(25; 1; 32), while our algorithm can be generalized to scale up non-conjugate Logistic Normal topic models.

We, next, plan to test out our inference algorithm for DTM in an even bigger scale with thousands of machines and terabytes of data. We intend to further extend this research and learn dynamic correlation graphs of topics over multiple time slices (Dynamic Correlated Topic Models), and improve the visualization of the learned evolution of topics from DTM, which has been hard to show graphically.

At the moment, we use Stochastic Gradient Langevin Dynamics on each node separately, but recent developments in Distributed Stochastic MCMC (2) can also parallellize the sampling of $\Phi_t$ and $\eta$ making the algorithm even more efficient.

Using SGLD requires tuning the step-size parameters, and the parameters can differ from dataset to dataset. Hence, we intend to try a more adaptive approach such as AdaGrad (11), or use more recent and sophisticated Stochastic Gradi-ent Monte Carlo methods such as SGHMC (9) and SGNHT (10) for inference.

## 7. CONCLUSIONS

We propose a scalable and efficient inference method of Dynamic Topic Models. Our algorithm is a novel combination of Stochastic Gradient Langevin Dynamics and Metropolis-Hastings sampler using Alias tables in a blockwise Gibbs Sampling framework. This combination makes our algorithm naturally parallelizable and allows it to scale up extremely well with the number of time slices and topics, as we have shown in our experiments.

Our algorithm is significantly faster than the baselines in both single machine and distributed environments. Making fewer restricting assumptions, our algorithm also performs slightly better in terms of perplexity than the existing variational methods. DTMs can capture very exciting topic trends over time but the existing inference methods use variational approximations and have not been able to learn large DTMs from big datasets.

Our algorithm has made it possible to do posterior inference of DTM at an industrial scale, and we prove this claim by learning the biggest existing DTM. Our work is applicable for both researchers and industries as a large scale Dynamic Topic Model can capture very interesting trends.

## Acknowledgments

The work was supported by the National Basic Research Program (973 Program) of China (Nos. 2013CB329403, 2012CB316301), National NSF of China (Nos. 61322308, 61332007), Tsinghua National Laboratory for Information Science and Technology Big Data Initiative, Tsinghua Initiative Scientific Research Program (No. 20141080934), and the Collaboration Awards from Microsoft and Intel.